\documentclass[11pt,a4paper]{article}
\usepackage[hyperref]{emnlp-ijcnlp-2019}
\usepackage{times}
\usepackage{latexsym}

\usepackage{url}

\usepackage{graphicx}
\usepackage{amsmath}
\usepackage{booktabs}
\usepackage{amssymb}
\usepackage{color}

\DeclareMathOperator*{\argmax}{argmax}

\aclfinalcopy 

\def\onedot{. }
\def\eg{\emph{e.g.},} 
\def\ie{\emph{i.e.},}

\def\etal{\emph{et al}\onedot}

\newcommand{\Tref}[1]{Table~\ref{#1}}
\newcommand{\Eref}[1]{Eq.~(\ref{#1})}
\newcommand{\Fref}[1]{Fig.~\ref{#1}}

\newcommand{\Cref}[1]{Chap.~\ref{#1}}

\title{Image Captioning with Very Scarce Supervised Data:\\ Adversarial Semi-Supervised Learning Approach
}

\author{%
  Dong-Jin Kim$^1$\quad%
  Jinsoo Choi$^1$\quad%
  Tae-Hyun Oh$^2$\quad%
  In So Kweon$^1$\\%
  $^1$KAIST, South Korea.\quad%
  $^2$MIT CSAIL, Cambridge, MA.\\
  {\footnotesize{$^1$\texttt{\{djnjusa,jinsc37,iskweon77\}@kaist.ac.kr } \quad $^2$\texttt{taehyun@csail.mit.edu} }}
  }

\date{}

\begin{document}

\maketitle
\begin{abstract}
{Constructing an organized dataset comprised of a large number of images and several captions for each image is a laborious task, which requires vast human effort.
On the other hand, collecting a large number of images and sentences separately may be immensely easier.}
In this paper, we develop a novel data-efficient \emph{semi-supervised} framework for training an image captioning model.
We leverage massive \emph{unpaired} image and caption data by learning to associate them. 
To this end, our proposed semi-supervised learning method assigns pseudo-labels to unpaired samples via Generative Adversarial Networks to learn the joint distribution of image and caption.
To evaluate, we construct {scarcely-paired COCO} dataset,
a modified version of MS COCO caption dataset.
The empirical results show the effectiveness of our method compared to several {strong} baselines, {especially when the amount of the paired samples are scarce}.
\end{abstract}

\section{Introduction}


Image captioning is a task of automatically generating a natural language description of a given image.
It is a highly useful task, in that 1) it extracts the essence
from an image into a self-descriptive form of representation, and 2) the output format is a natural language, which exhibits free-form and manageable characteristics useful to applications such as language based image or region retrieval~\cite{johnson2016densecap,karpathy2015deep,kim2019dense}, video summarization~\cite{choi2018contextually}, navigation~\cite{wang2018look}, vehicle control~\cite{kim2018textual}.
Image captioning exhibits free-form characteristics, since it is not confined to a few number of pre-defined classes.
This enables descriptive analysis of a given image.

Recent research on image captioning has made impressive progress~\cite{anderson2017bottom,vinyals2015show}.
Despite this progress, the majority of works are trained via supervised learning that requires a large corpus of caption labeled images such as the MS COCO caption dataset~\cite{lin2014microsoft}.
Specifically, the dataset is constructed with 1.2M images that were asked the annotators to provide five grammatically plausible sentences for each image.
Constructing such human-labeled datasets are an immensely laborious task and time-consuming.
This is a challenge of image captioning, because the task heavily requires large data, \ie~data hungry task.

\begin{figure}[t]
\begin{center}
   \includegraphics[width=1.0\linewidth]{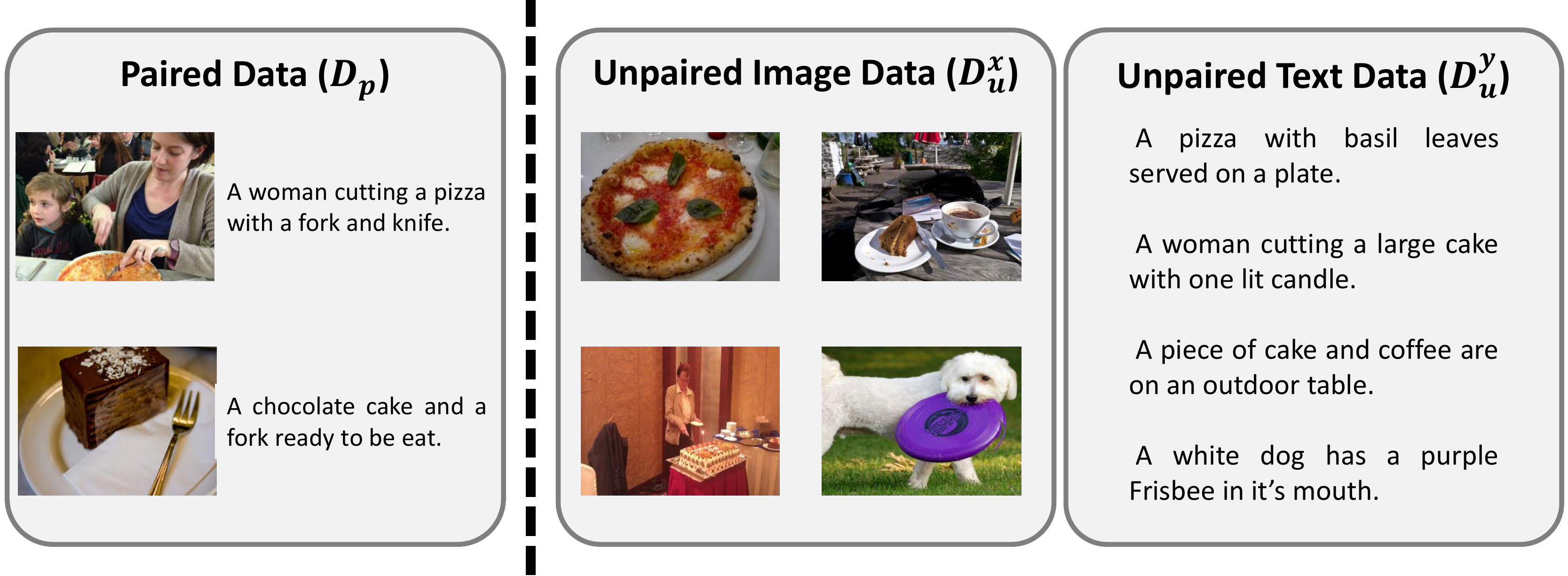}
   \vspace{-8mm}
\end{center}
   \caption{The proposed data setup utilizes both ``paired'' and ``unpaired'' image-caption data. We denote paired data as $\mathcal{D}_{p}$, and unpaired image and caption datasets as $\mathcal{D}_{u}^x$ and $\mathcal{D}_{u}^y$ respectively.
   \vspace{-2mm}
   }
\end{figure}

In this work, we present a novel way of leveraging \emph{unpaired} image and caption data to train data hungry image captioning neural networks.
We consider a scenario in which we have a small scale paired image-caption data of a specific domain. 
We are motivated by the fact that images can be easily obtained from the web and captions can be easily augmented and synthesized by replacing or adding different words for given sentences as done in~\cite{zhang2015character}.
Moreover, given a sufficient amount of descriptive captions, it is easy to crawl \emph{corresponding but noisy} images through Google or Flickr image databases ~\cite{thomee2016yfcc100m} to build an image corpus.
In this way, we can easily construct a scalable \emph{unpaired} dataset of images and captions, which requires no (or at least minimal) human effort.

Due to the unpaired nature of images (input) and captions (output supervision), the conventional supervision loss can no longer be used directly.
We propose to algorithmically assign supervision labels, termed as \emph{pseudo-labels}, to make unpaired data paired. 
The pseudo-label is used as a \emph{learned} supervision label.
To develop the mechanism of pseudo-labeling, we use generative adversarial network (GAN)~\cite{goodfellow2014generative}, for searching pseudo-labels for unpaired data.
That is, in order to find the appropriate pseudo-labels for unpaired samples, we utilize an adversarial training method for training a discriminator model.
Thereby, the discriminator learns to distinguish between real and fake image-caption pairs and to retrieve pseudo labels as well as to enrich the captioner training.
Our main contributions are summarized as follows. 
(1) We propose a novel framework for training an image captioner with the unpaired image-caption data and a small amount of paired data.
(2) In order to facilitate training with unpaired data, we devise a new semi-supervised learning approach by the novel usage of the GAN discriminator.
(3) 
We construct our \emph{scarcely-paired COCO} dataset, which is the modified version of the MS COCO dataset without pairing information. 
On this dataset, we show the effectiveness of our method in various challenging setups, compared to strong competing methods. 

\section{Related Work}
{The goal of our work is to deal with unpaired image-caption data for image captioning.
Therefore, we mainly focus on image captioning and unpaired data handling literature. 
}

\noindent \textbf{Data Issues in Image Captioning.}
Since the introduction of large-scale datasets such as MS COCO data~\cite{lin2014microsoft}, image captioning has been extensively studied in vision and language society~\cite{anderson2017bottom,rennie2017self,vinyals2015show,xu2015show} 
by virtue of the advancement of deep convolutional neural networks~\cite{krizhevsky2012imagenet}.
As neural network architectures become more advanced, they require much larger dataset scale for generalization~\cite{shalev2014understanding}.
Despite the extensive study on better network architectures, the data issues for image captioning, such as noisy data, partially missing data, and unpaired data have been relatively barely studied.

Unpaired image-caption data has been just recently discussed.
Gu~\etal\shortcite{gu2018unpaired}
introduce a third modal information, Chinese captions, 
for language pivoting~\cite{utiyama2007comparison,wu2007pivot}.
Feng~\etal\shortcite{feng2018unsupervised} propose an unpaired captioning framework which trains a model without image or sentence labels via learning a visual concept detector with external data, OpenImage dataset~\cite{krasin2017openimages}.
Chen~\etal\shortcite{chen2017show} approach image captioning as a domain adaptation by utilizing large paired MS COCO data as the source domain and adapting on a separate unpaired image or caption dataset
as the target domain.
Liu~\etal\shortcite{liu2018show} use self-retrieval for captioning to facilitate training a model with partially labeled data, where the self-retrieval module tries to retrieve corresponding images based on the generated captions.
As a separate line of work, there are novel object captioning methods~\cite{anne2016deep,venugopalan2017captioning} that {additionally} exploit unpaired image and caption data corresponding to a novel word.


Most of aforementioned works~\cite{gu2018unpaired,anne2016deep,venugopalan2017captioning,feng2018unsupervised} exploit large auxiliary supervised data which is often beyond image-caption data.
To the best of our knowledge, we are the first to study how to handle unpaired image and caption data for image captioning without any auxiliary information but by leveraging scarce paired image-caption data only.
Although \cite{chen2017show} does not use auxiliary information as well, 
it requires large amounts of paired source data, of which data regime is different from ours. 
\cite{liu2018show} is also this case, where they use the full paired MS COCO caption dataset with an additional large unlabeled image set.
Our method lies on a very scarce paired source data regime, of which scale is only $1\%$ of the COCO dataset. 


\noindent \textbf{Multi-modality in Unpaired Data Handling.}
By virtue of the advancement on generative modeling techniques, \eg~GAN~\cite{goodfellow2014generative}, multi-modal translation recently emerged as a popular field.
Among many possible modalities, image-to-image translation between two different (and unpaired) domains has been mostly explored.
To tackle this issue, the cycle-consistency constraint between unpaired data is exploited in CycleGAN~\cite{zhu2017unpaired} and DiscoGAN~\cite{kim2017learning}, and it is further improved in UNIT~\cite{liu2017unsupervised}.

In this work, we regard image captioning as a multi-modal translation.
Our work has similar motivation to the unpaired image-to-image translation~\cite{zhu2017unpaired}, and unsupervised machine translation~\cite{artetxe2018unsupervised,lample2018unsupervised,lample2018phrase} works {or machine translation with monolingual data~\cite{zhang2018joint}}.
However, we show that the cycle-consistency does not work on our problem setup due to a significant modality gap.
Instead, our results suggest that the traditional label propagation based semi-supervised framework~\cite{zhou2004learning} is more effective for our task. 





\noindent \textbf{Semi-supervised Learning.}
Our method is motivated by the generative model based semi-supervised learning~\cite{chongxuan2017triple,gan2017triangle,kingma2014semi}, which mostly deals with classification labels.
In contrast, we leverage caption data and devise a novel model to train the captioning model.

\section{Proposed Framework}


\begin{figure*}[t]
\begin{center}
   \includegraphics[width=.8\linewidth]{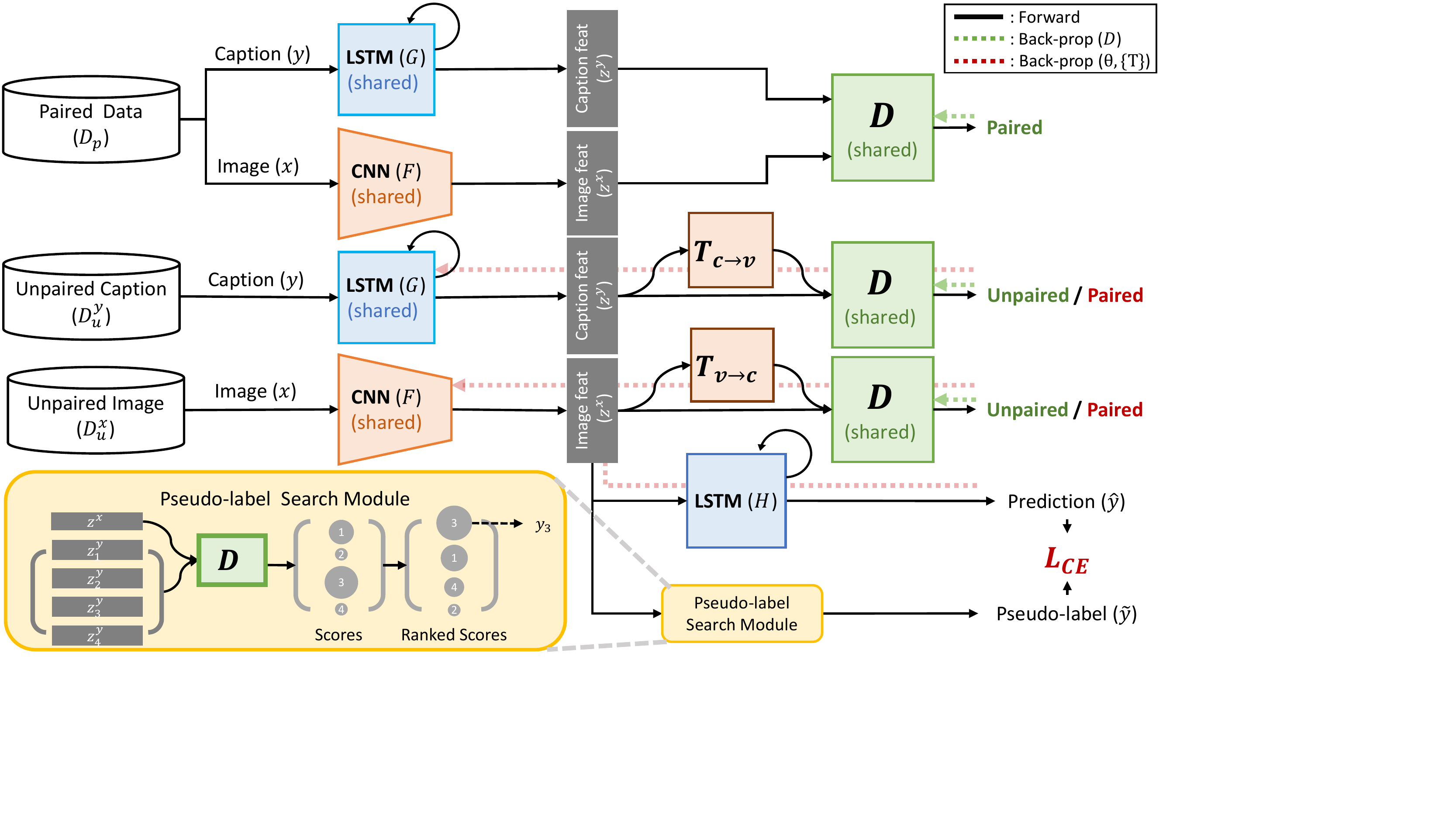}
\end{center}
   \vspace{-2mm}
   \caption{Description of the proposed method. Dotted arrows denote the path of the gradients via back-propagation. Given any image and caption pair, CNN and RNN (LSTM) encoders encode input image and caption into the respective feature spaces. A discriminator ($D$) is trained to discriminate whether the given feature pairs are real or fake, while the encoders are trained to fool the discriminator. 
   The learned discriminator is also used to assign the most likely pseudo-labels to unpaired samples through the pseudo-label search module.
   }
   \vspace{0mm}
   \label{fig:architecture}
\end{figure*}

In this section, 
we first brief the standard image caption learning, and describe how we can leverage the unpaired dataset.
Then, we introduce an adversarial learning method
for obtaining a GAN model that is used for assigning pseudo-labels and encouraging to match the distribution of latent features of images and captions.
Lastly, we describe a connection to CycleGAN.

\subsection{Model}
Let us denote a dataset with $N_p$ image-caption pairs as $\mathcal{D}_{p} = \{(\mathbf{x}_i, y_i)\}^{N_p}_{i=1}$. 
A typical image captioning task is defined as follows: 
given an image $\textbf{x}_i$, we want a model to generate a caption $y_i$ that best describe the image. 
Traditionally,
a captioning model is trained on a large paired dataset $(\mathbf{x},y)\in \mathcal{D}_{p}$, \eg~MS COCO dataset, 
by minimizing the negative log likelihood against the ground truth caption as follows:
\begin{equation}
\centering
    \min_\theta \sum^{}_{(\mathbf{x}, y) \in \mathcal{D}_{p}} L_\textsf{CE}(y,\mathbf{x}; \theta),
\label{eqn:cross-entropy}
\end{equation}
where $L_\textsf{CE} =-{\log q_{\theta}(y|\textbf{x})}$ denotes the cross entropy loss, $\theta$  the set of learnable parameters, and $q(\cdot)$ the probability output of the model.
Motivated by encoder-decoder based neural machine translation literature~\cite{cho2014learning}, traditional captioning frameworks are typically implemented as a encoder-decoder architecture~\cite{vinyals2015show}, \ie~CNN-RNN.
The CNN encoder $F(\mathbf{x};\theta_{enc})$ produces a latent feature vector $\mathbf{z}^x$ from a given input image $\mathbf{x}$, and the RNN decoder $H(\mathbf{z}^x;\theta_{dec})$ is followed to generate a caption $y$ in a natural language form from $\mathbf{z}^x$, as depicted in \Fref{fig:architecture}.
The term $q(\cdot)$ in \Eref{eqn:cross-entropy} is typically implemented by the sum of the cross entropy of each word token.
For simplicity, we omit $\theta$ from here if deducible, \eg~$H(\mathbf{z}^x)$.

\paragraph{Learning with Unpaired Data.}
Our problem deals with unpaired data samples, where the image and caption sets $\mathcal{D}_u^x {=} \{\mathbf{x}_i\}^{N_x}_{i=0}$ and $\mathcal{D}_u^y{=}\{y_i\}^{N_y}_{i=0}$ are not paired.
Given the unpaired datasets, 
due to the missing supervision, the loss in \Eref{eqn:cross-entropy} cannot be directly computed.
Motivated by the nearest neighbor aware semi-supervised framework~\cite{shi2018transductive},
we artificially generate \emph{pseudo-labels} for respective unpaired datasets.

Specifically, we retrieve a best match caption $\tilde{y}_i$ in $\mathcal{D}_u^y$ given a query image $\mathbf{x}_i$, and assign it as a pseudo-label, and vice versa ($\tilde{\mathbf{x}}_i$ for $y_i$).
We abuse the pseudo-label as a function for simplicity, \eg~$\tilde{y}_i = \tilde{y}(\mathbf{x}_i)$.
To retrieve a semantically meaningful match, we need a measure to assess matches.
We use a discriminator network to determine real or fake pairs by GAN, which we will describe later.
With the retrieved pseudo-labels, now we can compute \Eref{eqn:cross-entropy} by modifying it as:
\begin{equation}
\centering
\small
    \min_\theta
    \lambda_x\sum^{}_{\mathbf{x} \in \mathcal{D}_{u}^x} 
    L_\textsf{CE}(\tilde{y}(\mathbf{x}),\mathbf{x}; \theta) + 
    \lambda_y\sum^{}_{y \in \mathcal{D}_{u}^y} 
    L_\textsf{CE}(y,\tilde{\mathbf{x}}(y); \theta),
\label{eqn:cross-entropy2}
\end{equation}
where $\lambda_{\{\cdot\}}$ denote the balance parameters.


\paragraph{Discriminator Learning by Unpaired Feature Matching.}
We train the criterion to find a semantically meaningful match, so that pseudo-labels for each modality are effectively retrieved.
We learn a discriminator for that by using a small amount of paired supervised dataset.

We adopt a caption encoder, $G(y;\theta_{cap})$, which embeds the caption $y$ into a feature $\mathbf{z}^y$.
This is implemented with a single layer LSTM, and we take the output of the last time step as the caption representation $\mathbf{z}^y$.
Likewise, 
given an image $x$, we obtain $\mathbf{z}^x$ by
the image encoder $F(\mathbf{x}; \theta_{enc})$.
Now, we have a comparable feature space of $\mathbf{z}^x$ and $\mathbf{z}^y$.
We utilize the discriminator to distinguish whether the pair $(\mathbf{z}^x, \mathbf{z}^y)$ from true paired data $(\mathbf{x}, y)\in \mathcal{D}_p$, \ie~the pair belongs to the real distribution $p(\mathbf{x},y)$ or not.
We could use random pairs of $(\mathbf{x},y)$ independently sampled from respective unpaired datasets, but we found that
it is detrimental due to uninformative pairs.
Instead, we conditionally synthesize $\mathbf{z}^x$ or $\mathbf{z}^y$, {to form a} synthesized pair that appears to be as realistic as possible.
We use the feature transformer networks $\tilde{\mathbf{z}}^y\,{=}\,T_{v\rightarrow c}(\mathbf{z}^x)$ and $\tilde{\mathbf{z}}^x\,{=}\,T_{c\rightarrow v}(\mathbf{z}^y)$, where $v{\rightarrow}c$ denotes the mapping from visual data to caption data and vice versa, and  $\tilde{\mathbf{z}}$ denotes the conditionally synthesized feature.
$\{T\}$ are implemented with multi-layer-perceptron with four FC layers with ReLU nonlinearity.

The discriminator $D(\cdot,\cdot)$ learns to distinguish features, real or not. 
At the same time, the other associated networks $F$, $G$, $T_{\{\cdot\}}$ are learned to fool the discriminator by matching the distribution of paired and unpaired data.
Motivated by \cite{chongxuan2017triple}, the formulation of this adversarial training is as follows: 
\begin{equation}
\centering\footnotesize
\begin{aligned}
    \min_{\theta, \{T\}} 
    \max_{D} 
    & \mathop{\mathbb{E}}\limits_{\substack{(\mathbf{z}^x, \mathbf{z}^y) \\ \sim (F, G)\circ \mathcal{D}_p}}[\log(D(\mathbf{z}^x,\mathbf{z}^y)) + L_{reg}(\mathbf{z}^x, \mathbf{z}^y, \{T\})] \\
    &+ \tfrac{1}{2}\left( \mathop{\mathbb{E}}\limits_{\mathbf{x} \sim p (\mathbf{x})}[\log(1-D(F(\mathbf{x}),T_{v \rightarrow c}(F(\mathbf{x}))))]\right. \\
    &\left. + \mathop{\mathbb{E}}\limits_{y\sim p (y)}[\log(1-D(T_{c \rightarrow v}(G(y)),G(y)))] \right),
\end{aligned}\label{eqn:gan}
\end{equation}
\resizebox{\linewidth}{!}{$
\begin{aligned}
\textrm{where}\quad & L_{reg}(\mathbf{z}^x, \mathbf{z}^y, \{T\}){=}\\
&\lambda_{reg}( \|\mathop{T}\limits_{v \rightarrow
c}(\mathbf{z}^x) {-} \mathbf{z}^y\|_F^2 + \|\mathbf{z}^x {-} \mathop{T}\limits_{c \rightarrow v}(\mathbf{z}^y)\|_F^2), \nonumber
\end{aligned}$
}
and we use the distribution notation $p(\cdot)$ to flexibly refer to any type of the dataset regardless of $\mathcal{D}_p$ and $\mathcal{D}_u$.
Note that the first log term is not used for updating any learnable parameters related to $\theta,\{T\}$, but only used for updating $D$.
Through alternating training of the discriminator ($D$) and generator ($\theta,\{T\}$) similar to \cite{kim2018disjoint}, the latent feature distribution of paired and unpaired data should be close to each other, \ie~$p(\mathbf{z}^x,\mathbf{z}^y)=p(\mathbf{z}^x,T_{v\rightarrow c}(\mathbf{z}^x))=p(T_{c\rightarrow v}(\mathbf{z}^y),\mathbf{z}^y)$ (refer to the proof in \cite{chongxuan2017triple}). 
In addition, as the generator is trained, the decision boundary of the discriminator tightens. 
If the unpaired datasets are sufficiently large such that semantically meaningful matches exist between the different modality datasets, and if the discriminator $D$ is plausibly learned, 
we can use $D$ to retrieve a proper pseudo-label.
Our architecture is illustrated in \Fref{fig:architecture}.



\paragraph{Pseudo-label Assignment.}
Given an image $\mathbf{x}\in \mathcal{D}_u^x$, we retrieve a caption in the unpaired dataset, \ie~$\tilde{y} \in \mathcal{D}_u^y$, that has the highest score obtained by the discriminator, \ie~the most likely caption to be paired with the given image as
\begin{equation}
\centering
    \tilde{y}_i = \tilde{y}(\mathbf{x}_i) = \argmax_{y\in \mathcal{D}_u^y}\,\, D\left(F(\mathbf{x}_i), G(y) \right),
    \label{eqn:pseudo_x}
\end{equation}
vice versa for unpaired captions:
\begin{equation}
\centering
    \tilde{\mathbf{x}}_i = \tilde{\mathbf{x}}(y_i) = \argmax_{\mathbf{x}\in \mathcal{D}_u^x}\,\, D\left(F(\mathbf{x}), G(y_i) \right).
    \label{eqn:pseudo_y}
\end{equation}

By this retrieval process over all the unpaired dataset, 
we have image-caption pairs $\{(x_i, y_i)\}$ from the paired data and the pairs with pseudo-labels {$\{(x_j, \tilde{y}_j)\}$ and $\{(\tilde{x}_k, y_k)\}$} from the unpaired data. 
However, these pseudo-labels are not noise-free, 
thus treating them equally with the paired ones 
is detrimental.
Motivated by learning with noisy labels~\cite{lee2017cleannet,wang2018iterative}, we define a confidence score for each of the assigned pseudo-labels.
We use the output score from the discriminator, as the confidence score, \ie~$\alpha^x_i {=} \hat{D}(\mathbf{x}_i,\tilde{y}_i)$ and $\alpha^y_i {=} \hat{D}(\tilde{\mathbf{x}}_i,{y}_i)$, where we denote $\hat{D}(\mathbf{x}, y) \,{=}\, D(F(\mathbf{x}), G(y))$, and $\alpha\in [0,1]$ due to the sigmoid function of the final layer of the discriminator.
{Therefore, we utilize the confidence scores to assign weights to the unpaired samples.}
We compute the weighted loss as follows:\vspace{2mm}

\noindent
\resizebox{!}{2.5mm}{
$\min\limits_\theta \sum\limits_{(\mathbf{x},y) \in \mathcal{D}_{p}} 
    L_\textsf{CE}(y,\mathbf{x}; \theta) +
    \lambda_x\sum\limits_{\mathbf{x} \in \mathcal{D}_{u}^x} \alpha_i^x 
    L_\textsf{CE}(\tilde{y}(\mathbf{x}),\mathbf{x}; \theta)$
}
\begin{equation}
\resizebox{!}{2.5mm}{
    $+\lambda_y\sum\limits_{y \in \mathcal{D}_{u}^y} \alpha_i^y 
    L_\textsf{CE}(y,\tilde{\mathbf{x}}(y); \theta).$
}
\label{eqn:pseudo_cross_entropy}    
\end{equation}

We jointly train the model on both paired and unpaired data. 
To ease the training further, we add an additional triplet loss function:
\begin{equation}
    \sum_{\substack{(x_p,y_p)\in \mathcal{D}_{p},\\
    x_u\in \mathcal{D}_{u}^x,
    y_u\in \mathcal{D}_{u}^y}}
    \log\frac{p_{\theta}(y_p|x_p)}{p_{\theta}(y_p|x_u)}+\log\frac{p_{\theta}(y_p|x_p)}{p_{\theta}(y_u|x_p)},
\label{eqn:triplet}
\end{equation}
by regarding random unpaired samples as negative.
This slightly improves the performance.

\subsection{Connection with CycleGAN}
As a representative fully unpaired method, 
CycleGAN~\cite{zhu2017unpaired} would be the strong baseline at this point,
which has been popularly used for unpaired distribution matching.
Since it is designed for image-to-image translation, we describe the modifications to it to fit our task, so as to understand relative performance of our method over the CycleGAN baseline.
When applying the cycle-consistency on matching between images and captions, 
since the input modalities are totally different, we modify it to a translation problem over feature spaces as follows:
\begin{equation}
\centering\small
\begin{aligned}
    &\min_{\theta, \{T\}} \max_{D_{\{x,y\}}}  L_{cycle}(\{T\})\\ 
    &+\mathop{\mathbb{E}}\limits_{\mathbf{x}\sim p(\mathbf{x})}[\log(D_x(F(\mathbf{x}))) + \log(1-D_x(T_{v\rightarrow c}(F(\mathbf{x}))))]\\
    &+\mathop{\mathbb{E}}\limits_{y\sim p(y)}[\log(D_y(G(y))) + \log(1-D_y(T_{c\rightarrow v}(G(y))))],
\end{aligned}
\label{eqn:CycleGAN}
\end{equation}
where 
{$\{T\}$ and $D_{\{x,y\}}$ denotes the feature translators and the discriminators for image domain and caption domain, and }
\begin{equation}
\centering\small
\begin{aligned}
	L_{cycle}(\{T\}) = \mathbb{E}_{\mathbf{x}\sim p(\mathbf{x})}[\lVert T_{c\rightarrow v}(T_{v\rightarrow c}(F(\mathbf{x})))-F(\mathbf{x})\rVert_2]  
    \\+ \mathbb{E}_{y\sim p(y)}[\lVert T_{v\rightarrow c}(T_{c\rightarrow v}(G(y)))-G(y)\rVert_2].
\end{aligned}
\label{eqn:Cycle-Consistency}
\end{equation}
The discriminator $D$ is learned to distinguish whether the latent features is from the image or the caption distribution. 
This is different with our method, in that we distinguish the correct \emph{semantic match of pairs}.
We experimented with CycleGAN purely on unpaired datasets, but we were not able to train it; hence, we add the supervised loss (\Eref{eqn:cross-entropy}) with the paired data, which is a fair setting with ours.
\section{Experiments}

In this section, we describe the experimental setups, competing methods and provide performance evaluations of unpaired captioning with both quantitative and qualitative results.

\subsection{Implementation Details}
We implement our neural networks by PyTorch library~\cite{paszke2017automatic}.
We use ResNet101~\cite{he2016deep} model as a backbone CNN encoder for image input, and initialize with weights pre-trained on ImageNet~\cite{russakovsky2015imagenet}.
Each region is represented as 2,048-dimensional output from the last residual block of this network.
For NIC model~\cite{vinyals2015show} and Att2in model~\cite{rennie2017self} in \Tref{table:captioning_unpaired}, we use visual features with size $2048 \times 7 \times 7$ .
For Up-Down~\cite{anderson2017bottom} and the model in \Tref{table:captioning_unpaired}, we use Faster R-CNN~\cite{ren2015faster} pre-trained on Visual Genome~\cite{krishna2017visual} object dataset, thereby we extract visual features for 10 to 100 regions.

We set the {channel} size of the hidden layers of LSTMs to be 1024, 512 for the attention layer, and 1024 for the word embeddings.
For inference stage, we empirically choose the beam size to be 3 when generating a description, which shows the best performance.

We use a {mini}batch size of 100, the Adam~\cite{ba2015adam} optimizer for training (learning rate $lr {=} 5e^{-4}$, $b_1{=}0.9$, $b_2{=}0.999$).
For hyper-parameters, we empirically choose  $\lambda_x$ and $\lambda_y$ to be equal to 0.1.
The total loss function for training our model is as follows:
\begin{equation}
\mathcal{L} = \mathcal{L}_{cap} + 0.1 \mathcal{L}_{GAN} + 0.1 \mathcal{L}_{triplet},
\end{equation}
where $\mathcal{L}_{cap}$ denotes the captioning loss defined in \Eref{eqn:pseudo_cross_entropy}, $\mathcal{L}_{GAN}$ the loss for adversarial training defined in \Eref{eqn:gan}, and $\mathcal{L}_{triplet}$  the triplet loss defined in \Eref{eqn:triplet}.

\subsection{Experimental Setups}
\label{setup}

\begin{table}[t]
\centering
    \resizebox{1.0\linewidth}{!}{%
		\begin{tabular}{l|| ccc|cccc}\hline
			&		(A)	&(B)&	(C)		&BLEU1&BLEU4&METEOR&CIDEr	\\\hline\hline
			Fully paired (100\%)  &		 	&	 		&	 		&72.5 &	29.4 &25.0&95.7		\\
			\hline
			\hline
			Paired only (1\%)	                &		     &	          &		      &58.1 &   13.4	 &15.9&36.0		\\
			CycleGAN (Zhu~\etal) &            &	          &		       &58.7&	14.1 &16.2&37.7		\\
			\textbf{Ours ver1}              &$\checkmark$&            &		       &60.1&	15.7 &16.5&45.3		\\
			\textbf{Ours ver2}              &$\checkmark$&$\checkmark$&		       &61.3&	17.1 &19.1&51.7		\\
			\textbf{Ours (final)}           &$\checkmark$&$\checkmark$&$\checkmark$&\textbf{63.0}&	\textbf{18.7} &\textbf{20.7}&\textbf{55.2}		\\
			\hline
			\hline
			\cite{gu2018unpaired} &&& &46.2& 5.4 &   13.2    &   17.7\\
			\cite{feng2018unsupervised} &&& &58.9& 18.6 &   17.9    &   54.9\\
			\hline
		\end{tabular}
    }
    \vspace{-1mm}
	\caption{Captioning performance comparison on MS COCO testset.
	The ``Paired only'' baseline is trained only with 1\% of paired data from our \emph{scarcely-paired COCO} dataset, while CycleGAN and Ours \{ver1, ver2, final\}, are trained with our \emph{scarcely-paired COCO} dataset (1\% of paired data and unpaired image and caption datasets).
	We indicate the ablation study by: (A) the usage of the proposed GAN that distinguishes real or fake image-caption pairs, (B) pseudo-labeling, and (C) noise handling by sample re-weighting.
    We also compare with \cite{gu2018unpaired} and \cite{feng2018unsupervised}, which are trained with unpaired datasets.
	\vspace{-0mm}
	}
	\label{table:ablation}	
\end{table}

We utilize MS COCO caption~\cite{lin2014microsoft} as our target dataset, which contains 123k images 
with 5 caption labels per image. 
To validate our model, we follow \emph{Karpathy} splits~\cite{karpathy2015deep}, which have been broadly used in image captioning literature.
The Karpathy splits contain 113k training, 5k validation, and 5k test images in total.
In our experiments, in order to simulate the scenario that both paired and unpaired data exist, we use two different data setups: 1) the proposed scarcely-paired COCO, and 2) partially labeled COCO~\cite{liu2018show} setup.

\begin{figure}
\vspace{0mm}
\begin{center}
   \includegraphics[width=0.7\linewidth]{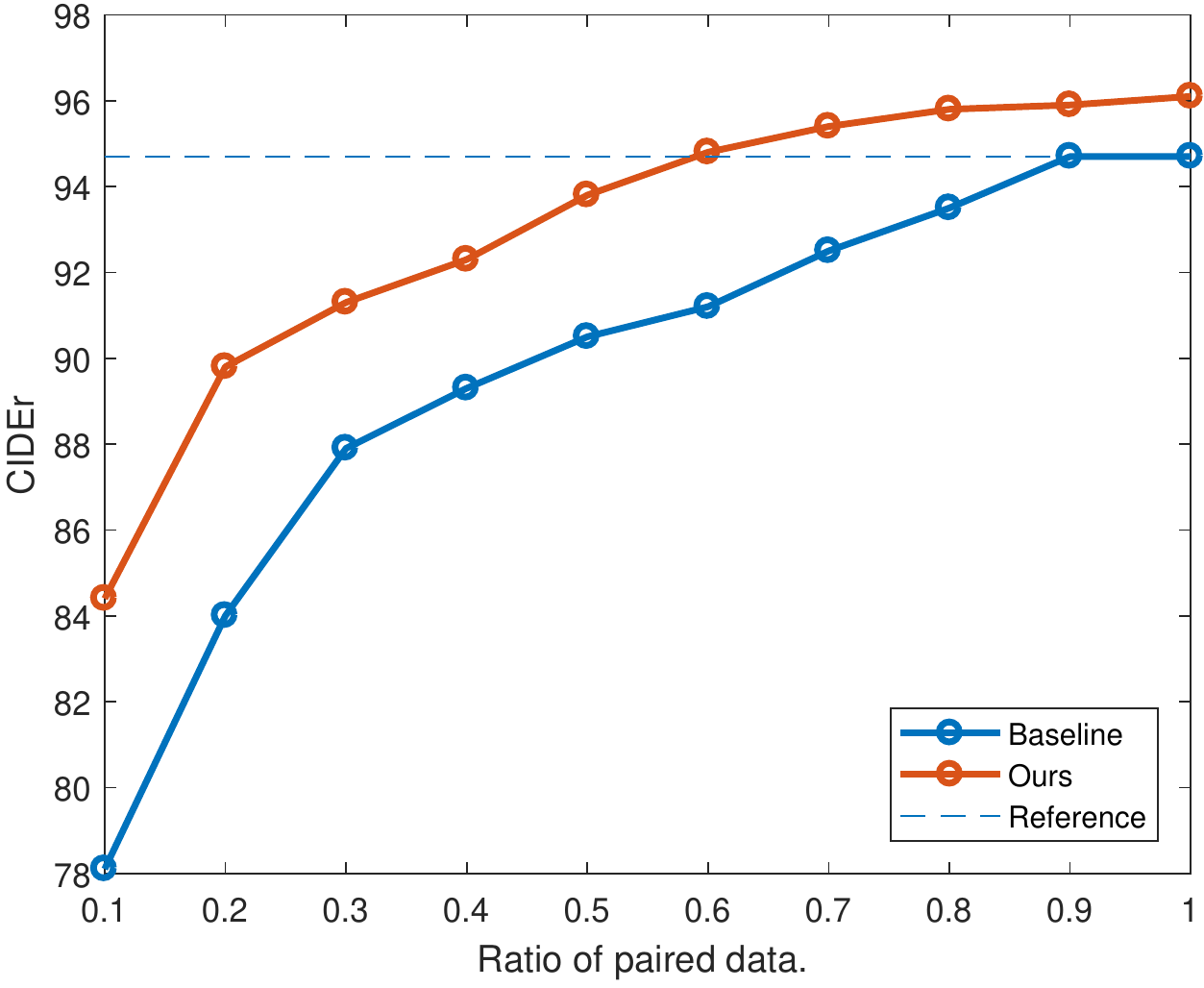}
\end{center}
    \vspace{-1mm}
   \caption{Performance w.r.t. amount of paired data for training. 
   Baseline denotes our \emph{Paired Only} baseline, Ours is our final model, and Reference is \emph{Paired Only} trained with full paired data.
   \vspace{-1mm}}
\label{fig:paired}
\end{figure}

\begin{figure*}[!ht]
\begin{center}
   \includegraphics[width=1.0\linewidth]{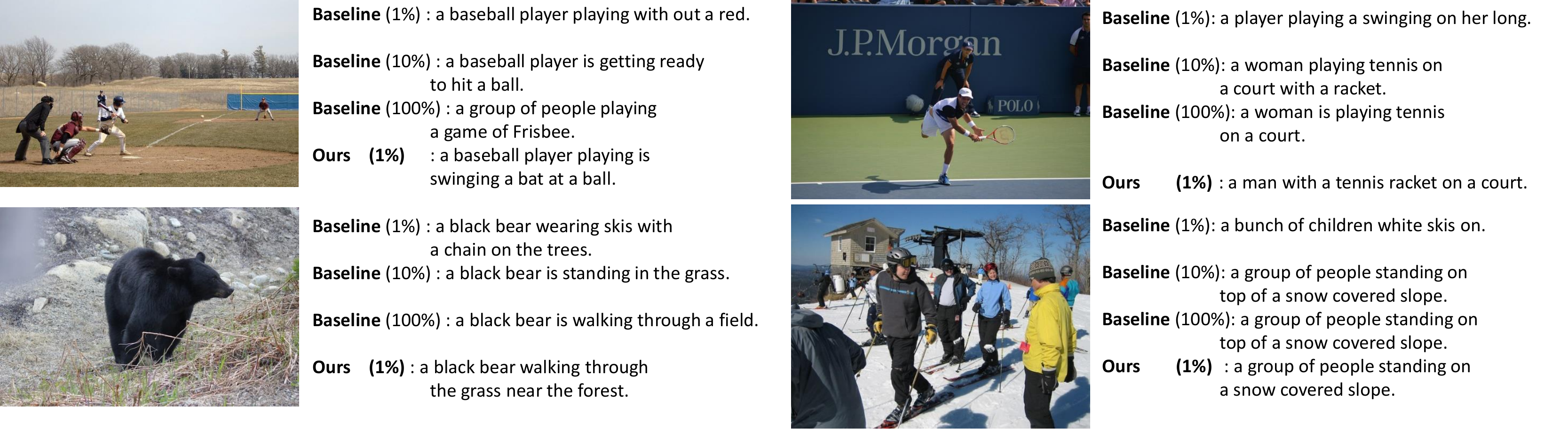}\\
   \includegraphics[width=1\linewidth]{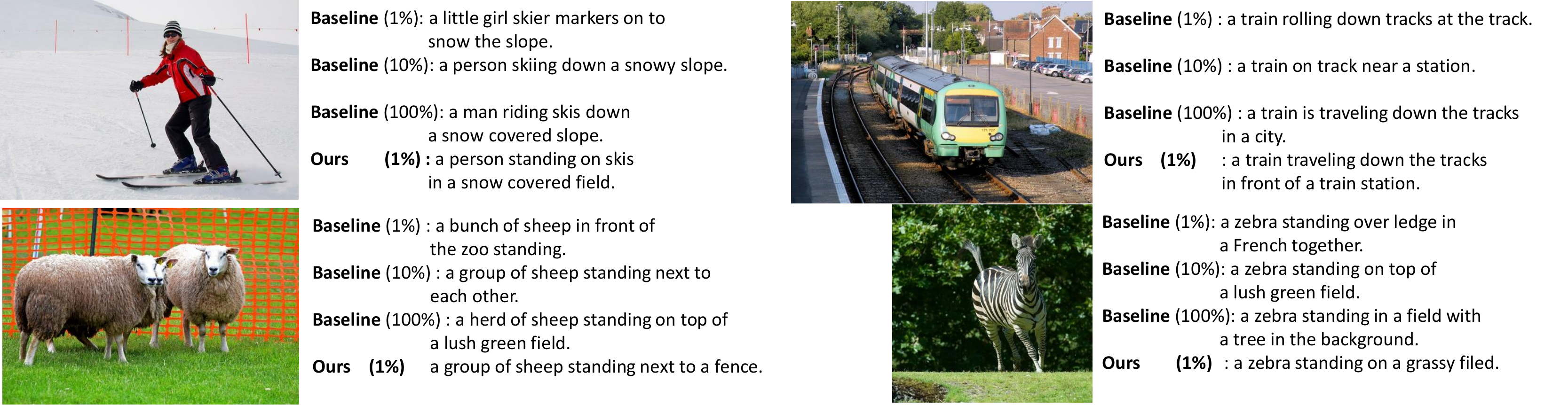}
\end{center}
    \vspace{-1mm}
   \caption{Sampled qualitative results of our model.
   We compare with the baseline models trained only on $N\%$ of paired samples out of the full MS COCO.
   Despite the use of only 1\% paired data, our model
   generates plausible captions similar to that of the baseline models trained with more data (10\% and above).   }
   \vspace{-0mm}
\label{fig:qualitative}
\end{figure*}

For the \emph{scarcely-paired} COCO setup, we remove the pairing information of the MS COCO dataset, while leaving a small fraction of pairs unaltered.
We randomly select only one percent of the total data, \ie~1,133 training images for the paired data, and remove the pairing information of the rest to obtain unpaired data. 
We call the small fraction of samples given as pairs as \emph{paired} data ($\mathcal{D}_p$) and call the samples without pairing information as \emph{unpaired} data ($\mathcal{D}_u$). 
We study the effects of other ratios of paired data used for training (in Figs.~\ref{fig:qualitative}~and~\ref{fig:paired}).
This dataset allows evaluating the proposed framework to measure whether such small paired data can be leveraged to learn plausible pseudo-label assignment, and what performance can be achieved compared to the fully supervised case.



For \emph{partially labeled} COCO setup, we follow \cite{liu2018show} and use the whole MS COCO data (paired) and add the \emph{``Unlabeled-COCO''} split of the officially MS-COCO~\cite{lin2014microsoft} for unpaired images, which involves 123k images without any caption label.
To compute the cross entropy loss, 
we use the pseudo-label assignment for the Unlabeled-COCO images.

For evaluation, we use the following metrics conventionally used in image captioning: BLEU~\cite{papineni2002bleu}, ROUGUE-L~\cite{lin2004rouge}, SPICE~\cite{anderson2016spice}, METEOR~\cite{denkowski2014meteor}, and CIDEr~\cite{vedantam2015cider}.  
All the evaluation is done on MS COCO testset.

In order to avoid high time complexity of the pseudo-labeling process, we do not search the pseudo-labels from the whole unpaired data.
Pseudo-label retrieval is done on a subset of one-hundredth of the unpaired data (yielding 1000 samples). 
Thus, the complexity of each minibatch becomes $O(B\times N \times 0.01),$ where $B$ denotes the minibatch size, $N$ denotes the size of the unpaired dataset.
Since we also apply label-noise handling, a plausible pseudo-label assignment is sufficient in helping the learning process. 
We can improve performance by using a larger subset or by using a fast approximate nearest neighbor search.

\subsection{Evaluation on Scarcely-paired COCO}
We follow the same setting with \cite{vinyals2015show}, if not mentioned. 
\Tref{table:ablation} shows the results on 
the {scarcely-paired COCO} dataset. 
We compare with several baselines: 
\emph{Paired Only}; we train our model only on the small fraction (1\%) of the paired data, \emph{CycleGAN}; 
we train our model with the cycle-consistency loss~\cite{zhu2017unpaired} (as in \Eref{eqn:CycleGAN}).
Additionally, we train variants of our model denoted as \emph{Ours} (ver1, ver2, and final). 
The model denoted as \emph{Our ver1}, is trained via our GAN model (\Eref{eqn:gan}) that distinguishes real or fake image-caption pairs.
For \emph{Our ver2}, we add training with pseudo-labels for unpaired data with  \Eref{eqn:pseudo_cross_entropy} while setting the confidence scores $\alpha^x{=}\alpha^y{=}1$ for all training samples.
For our final model denoted as \emph{Our (final)}, we apply the noise handling technique by the confidence scores {$\alpha^x$ and $\alpha^y$} computed by \Eref{eqn:pseudo_cross_entropy} and re-weighting the loss for each sample.
We present the performance of the fully supervised (\emph{Fully paired}) model using 100\% of the COCO training data for reference.

\begin{table*}[t]
\centering
    \resizebox{.9\linewidth}{!}{%
		\begin{tabular}{l|| cccccccccc}\hline
			&BLEU1	&	BLEU2		&BLEU3	&BLEU4&ROUGE-L&SPICE&METEOR&CIDEr	\\\hline\hline
			Self-Retrieval~\cite{liu2018show} (w/o unlabeled) 	&		79.8	&	62.3		&	47.1	&34.9 &56.6&20.5&27.5&114.6		\\		
			Self-Retrieval~\cite{liu2018show} (with unlabeled)  &		80.1	&	63.1		&	48.0	&35.8 &57.0 &21.0 &27.4&117.1		\\
			\hline
			Baseline (w/o unlabeled)  &		77.7	&	61.6  &	46.9&	36.2 &  56.8    &20.0&26.7&114.2		\\
			\textbf{Ours} (w/o unlabeled) &		80.8    &	65.3  &	49.9&	37.6 &  58.7    &   22.7    &   28.4    &   122.6		\\
			\textbf{Ours} (with unlabeled)  &\textbf{81.2}&\textbf{66.0}&\textbf{50.9}&\textbf{39.1}&\textbf{59.4}&\textbf{23.8}&\textbf{29.4}&\textbf{125.5}\\
			\hline
		\end{tabular}
    }
    \vspace{-1mm}
	\caption{
	Comparison with the semi-supervised image captioning method, ``Self-Retrieval''~\protect\cite{liu2018show}.
	Our method shows improved performance even without Unlabeled-COCO data (denoted as \emph{w/o unlabeled}) as well as with Unlabeled-COCO (\emph{with unlabeled}), although our model is not originally proposed for such scenario.
	\vspace{-0mm}
	}
	\label{table:captioning_semi}	
\end{table*}

As shown in \Tref{table:ablation}, in a scarce data regime, utilizing the unpaired data improves the captioning performance in terms of all metrics by noticeable margins.
Also, our models show favorable performance compared to the CycleGAN model in all metrics.
Our final model with pseudo-labels and noise handling achieves the best performance in all metrics across all the competitor, hereafter refered to this model as our model.

We also compare the recent unpaired image captioning methods~\cite{gu2018unpaired,feng2018unsupervised} in  \Tref{table:ablation}.
Both of the methods are evaluated on MS COCO testset.
In the case of Gu~\emph{et al{.}}, AIC-ICC image-to-Chinese dataset~\cite{wu2017ai} is used as unpaired images $\mathcal{D}_u^x$ and captions from MS COCO are used as unpaired captions $\mathcal{D}_u^y$.
Note that this is not a fair comparison to our method but they have more advantages, in that 
Gu~\emph{et al{.}} use large amounts of additional labeled data (10M Chinese-English parallel sentences of AIC-MT dataset~\cite{wu2017ai}) and Feng~\emph{et al{.}} use 36M samples of additional OpenImages dataset, whereas our model only uses a small amount of paired samples (1k) and 122k unpaired data.
Despite far lower reliance on paired data, our model shows favorable performance against recent unpaired captioning works.


We study our final model against our \emph{Paired Only} baseline
according to varying amounts of paired training data
{in \Fref{fig:paired}}, so that we can see 
how much information can be gained from the unpaired data.
From 100\% to 10\%, as the amount of paired samples decreases, the fluency and the accuracy of the descriptions get worse.
In particular, we observed that most of the captions generated from the {\emph{Paired Only} baseline} trained with 10\% of paired data (11,329 pairs) show erroneous grammatical structure.
In contrast, by leveraging unpaired data, our method can generate more fluent and accurate captions, compared to \emph{Paired Only}  trained on the same amount of paired data. 
It is worthwhile to note that
our model trained with 60\% of paired data (67,972 pairs) achieves similar performance to the \emph{Paired Only} baseline trained with fully paired data (113,287 pairs), which signifies that our method is able to save \emph{near half} of the human labeling effort of constructing a dataset.

We also show qualitative samples of our results in \Fref{fig:qualitative}. 
\emph{Paired Only} baseline trained with 1\% paired data produces erroneous captions, and the baseline with 10\% paired data 
starts to produce plausible captions.
It is based on \emph{ten times} more paired samples, compared to our model that uses only 1\% of them.
We highlight that, in the two examples on the top row {of \Fref{fig:qualitative}}, our model generates more accurate captions than the {\emph{Paired Only}} baseline trained on the 100\% paired data (``baseball'' to ``Frisbee'' on the top-left, and ``man'' to ``woman'' on the top-right).
This suggests that unpaired data with our method effectively boosts the performance
especially when paired data is scarce.

\begin{figure}
\begin{center}
   \includegraphics[width=1.0\linewidth]{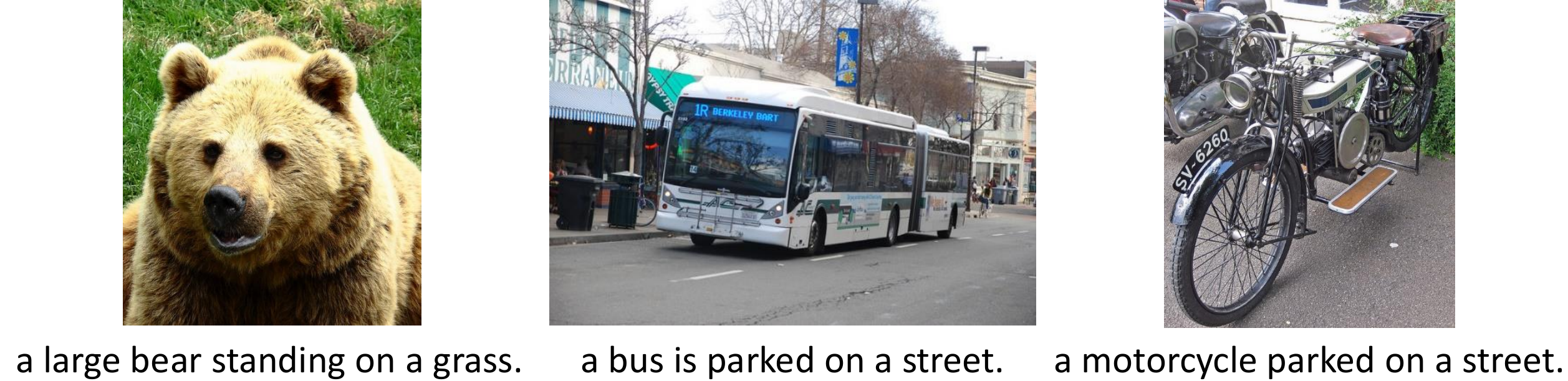}
\end{center}
   \vspace{-1mm}
   \caption{Examples of the pseudo-labels (captions) assigned to the unpaired images. Our model is able to sufficiently assign image-caption pairs through the proposed adversarial training. }
   \vspace{-1mm}
\label{fig:pseudo-label}
\end{figure}

In order to demonstrate the meaningfulness of our pseudo-label assignment, we show the pseudo-labels assigned to the unpaired samples.
We show the pseudo-labels (captions) assigned to unlabeled images from Unlabeled-COCO images in \Fref{fig:pseudo-label}.
As shown in the figure, despite not knowing the real pairs of these images, the pseudo labels are sufficiently assigned by the model.
Note that even though there is no ground truth caption for unlabeled images in the searching pool, the model is able to find the most likely (semantically correlated) image-caption pair for the given images.

\begin{figure}
\begin{center}
   \includegraphics[width=1\linewidth]{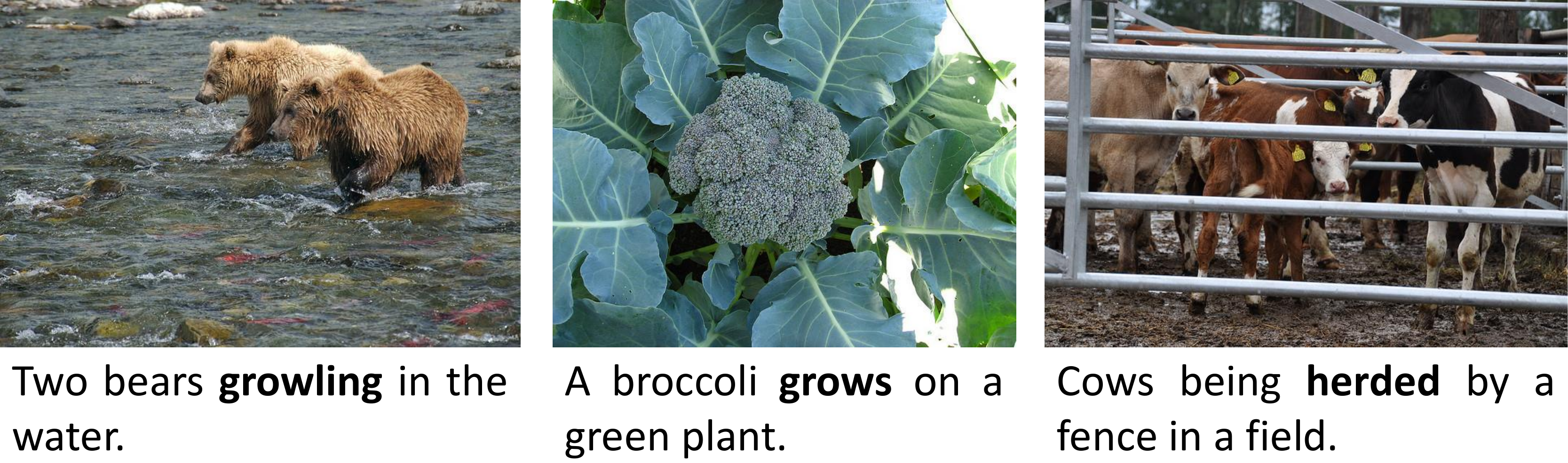}
\end{center}
\vspace{-2mm}
   \caption{Example captions containing words that do not exist in the paired dataset $\mathcal{D}_p$.
   The novel words that are not in $\mathcal{D}_p$ but in $\mathcal{D}_u^y$ are highlighted in bold.}
   \vspace{-1mm}
\label{fig:novel}
\end{figure}

\Fref{fig:novel} highlights interesting result samples of our caption generation, where
the results contain words that do not exist in the paired data of the scarcely-paired COCO dataset.
The semantic meaning of the words such as ``growling,'' ``growing,'' and ``herded,'' which involve in the unpaired caption data but not in the paired data, may have been learned properly via pseudo-label assignment during training.
These examples suggest that our method is capable to infer the semantic meaning of  unpaired words to some extent, which would have been unable to be learned with only little paired data. 
This would be an evidence that our framework is capable to align abstract semantic spaces between two modalities, \ie~visual and caption.

\begin{table*}[t]
\centering
    \vspace{0mm}
    \resizebox{.8\linewidth}{!}{%
		\begin{tabular}{l|| cccccccccc}\hline
			&BLEU1	&	BLEU2		&BLEU3	&BLEU4&ROUGE-L&SPICE&METEOR&CIDEr	\\\hline\hline
			NIC~\cite{vinyals2015show} 	&		72.5	&	55.1		&	40.4		 &	29.4   &52.7&18.2&25.0&95.7		\\
			NIC \textbf{+Ours}  &		\textbf{74.2}	&	\textbf{56.9}		&	\textbf{42.0}&	\textbf{31.0}            &\textbf{54.1}&\textbf{19.4}&\textbf{28.8}&\textbf{97.8}		\\
			\hline
			Att2in~\cite{rennie2017self} 	&		74.5	&	58.1		&	44.0		&	33.1 &54.6 & 19.4&26.2&104.3	\\	
			Att2in \textbf{+Ours}  &		\textbf{75.5}	&	\textbf{58.8}		&	\textbf{45.0}		&	\textbf{34.3} &\textbf{55.4}&\textbf{20.0}&\textbf{26.8}&\textbf{105.3}	\\
			\hline
			Up-Down~\cite{anderson2017bottom} 	&		76.7	&	60.7		&	47.1		&	36.4 &56.6&20.6&27.6&113.6		\\	
			Up-Down \textbf{+Ours}  &		\textbf{77.3}	& \textbf{64.1}		&	\textbf{47.6}		&	\textbf{37.0} &\textbf{57.1}&\textbf{21.0}&\textbf{27.8}&\textbf{114.6}		\\
			\hline
		\end{tabular}
    }
	\vspace{-1mm}
	\caption{Evaluation of our method with different backbone architectures.
	All models are reproduced and trained with the cross entropy loss.
	By adding Unlabeled-COCO images, our training method was applied in a semi-supervised way, which shows consistent improvement in all the metrics. 
	}
	\vspace{-0mm}
	\label{table:captioning_unpaired}	
\end{table*}

\subsection{Evaluation on Partially Labeled COCO}
\label{sec:semi}
For a more realistic setup, we compare with the recent semi-supervised image captioning method, called \emph{Self-Retrieval} \cite{liu2018show}, on their problem regime, \ie~``partially labeled''  COCO setup, where the full amount of the paired MS COCO (113k) plus 123k uncaptioned images of the Unlabeled-COCO set are used (no additional unpaired caption is used).
While their regime is not our direct scope, but we show the effectiveness of our method in the regime.
Then, we extend our framework by replacing our backbone architecture with recent advanced image caption architectures.
In this setup, a separate \emph{unpaired} caption data $\mathcal{D}_u^y$ does not exist, thus we use captions from paired COCO data $\mathcal{D}_p$ as pseudo-labels.

\Tref{table:captioning_semi} shows the comparison with Self-Retrieval.
For a fair comparison with them, we replace the cross entropy loss from our loss with the policy gradient method~\cite{rennie2017self} to directly optimize {our} model with CIDEr score as in \cite{liu2018show}.
{As our baseline model (denoted as \emph{Baseline}), we train a model only with policy gradient method without the proposed GAN model.}
When only using the 100\% paired MS COCO dataset (denoted as \emph{w/o unlabeled}), our model already shows improved performance over Self-Retrieval.
Moreover, when adding Unlabeled-COCO images {(denoted as \emph{with unlabeled})}, our model performs favorably against Self-Retrieval  in all the metrics.
The results suggest that our method is also advantageous in the semi-supervised setup.

To further validate our method in the semi-supervised setup, 
we compare different backbone architectures~\cite{anderson2017bottom,rennie2017self,vinyals2015show} in our framework, where their methods are developed for fully-supervised methods.
We use the same data setup with the above, but we replace CNN ($F$) and LSTM ($H$) from our framework with the image encoder and the caption decoder from their image captioning models. 
Then, these models can be trained in our framework as it is by alternating between the discriminator update and pseudo-labeling.
\Tref{table:captioning_unpaired} shows the comparison.
Training with the additional Unlabeled-COCO data via \emph{our} training scheme consistently improves all baselines in all metrics.



\subsection{Captioning with Web-Crawled Data}
To simulate a scenario involving crawled data from the web, we use the setup suggested by Feng~\etal\shortcite{feng2018unsupervised}.
They collect a sentence corpus by crawling the image descriptions from Shutterstock\footnote{https://www.shutterstock.com} as unpaired caption data $\mathcal{D}_u^y$, whereby 2.2M sentences are collected.
For unpaired image data $\mathcal{D}_u^x$, they use only the images from the MS COCO data, 
while the captions are not used for training.
For training our method, we leverage from 0.5\% to 1\% of paired MS COCO data as our paired data $\mathcal{D}_p$.
The results are shown in \Tref{table:shutterstock} with the performance obtained by Feng~\emph{et al}.
Note again that Feng~\emph{et al.} exploits external large-scale data, \ie~36M images in the OpenImages dataset.
With 0.5\% of paired only data (566 pairs), our baseline shows lower scores in terms of BLEU4 and METEOR than Feng~\emph{et al.}, while our proposed model shows comparable or favorable performance in BLEU4, ROUGE-L and METEOR.
Our method starts to have higher scores in all metrics from 0.8\% of paired data (906 pairs), even without additional 36M images.


\begin{table}[t]
\centering
    \resizebox{1.0\linewidth}{!}{%
		\begin{tabular}{l|| ccccc}\hline
			&		BLEU4&ROUGE-L&SPICE&METEOR&CIDEr	\\\hline\hline
			Paired only (0.5\% paired)	  &   4.4	&33.7   &3.6    &10.8   &8.6		\\
			\textbf{Ours} (0.5\% paired)  &5.4      &34.6   &4.2    &12.0   &10.5		\\
			\hline
			Paired only (0.7\% paired)	  &   3.5	 &36.1  &3.7    &11.4   &8.9		\\
			\textbf{Ours} (0.7\% paired)  &8.5      &39.0   &5.2    &13.6   &20.2		\\
			\hline
			Paired only (0.8\% paired)	  &   8.8	 &39.1  &5.9    &13.2   &21.9		\\
			\textbf{Ours} (0.8\% paired)  &   12.2	 &41.6  &7.6    &15.1   &29.0	\\
			\hline
			Paired only (1\% paired)	 &   13.4   &41.9   &8.3    &15.9   &36.0		\\
			\textbf{Ours}  (1\% paired)  &15.2      &43.3   &9.4    &16.9   &39.7		\\
			\hline
			\hline
			\cite{feng2018unsupervised} &4.8        &27.2   &7.0    &11.4   &23.3\\
			\hline
		\end{tabular}
    }
    \vspace{-1mm}
	\caption{Performance comparison with web-crawled data (Shutterstock).
    On top of unpaired image and caption data, our method is trained with 0.5 -- 1\% of paired data, while Feng \emph{et al.} use 36M additional images of the OpenImage dataset.
	\vspace{-1mm}
	}
	\label{table:shutterstock}	
\end{table}

\section{Conclusion}
We introduce a method to train an image captioning model with a large scale unpaired image and caption data, given a small amount of paired data. 
Our framework achieves favorable performance compared to various methods and setups. 
Unpaired captions and images are the data that can be easily collected from the web.
It can facilitate application specific captioning models, where labeled data is scarce. 

\vfill

\noindent{\textbf{Acknowledgements.}}
This work was supported by Institute for Information \& communications Technology Planning \& Evaluation(IITP) grant funded by the Korea government(MSIT) (No.2017-0-01780, The technology development for event recognition/relational reasoning and learning knowledge based system for video understanding)

\bibliography{egbib}
\bibliographystyle{named}



\end{document}